\documentclass{article}
\usepackage[dblblindworkshop,final]{neurips_2026}
\workshoptitle{Beyond Private Training: The New Landscape of AI Privacy (InfPriv)}
\usepackage{float}
\usepackage[utf8]{inputenc}
\usepackage[T1]{fontenc}
\usepackage{amsmath,amsfonts,graphicx}
\usepackage{hyperref,url,booktabs,nicefrac,microtype,xcolor}
\usepackage{newunicodechar}
\hypersetup{hidelinks} 

\newunicodechar{ı}{\i}
\newunicodechar{—}{\textemdash}
\newunicodechar{–}{\textendash}
\newunicodechar{’}{'}
\title{Constructing and Validating Memorisation Audits for EHR
Foundation Models}
\author{%
  Inigo Stibbe\textsuperscript{1} \quad
  Joshua Placidi\textsuperscript{1}\quad
  Padmanabhan Ramnarayan\textsuperscript{2} \quad
  A. Aldo Faisal\textsuperscript{1,3} \\[0.5em]
  \textsuperscript{1}Departments of Computing and Bioengineering,
  Imperial College London, UK \\
  \textsuperscript{2}Department of Surgery \& Cancer,
  Imperial College London, UK \\
  \textsuperscript{3}Chair in Digital Health \& Data Science,
  University of Bayreuth, Germany \\[0.5em]
  {\small\texttt{i.stibbe25@imperial.ac.uk}}
}

\begin{document}
\maketitle

\begin{abstract}
Autoregressive electronic health record (EHR) foundation models may disclose training information at inference, yet clinical predictability makes patient-specific memorisation difficult to distinguish from generalisation. We evaluate fixed-window token reconstruction, sensitive-attribute disclosure, and random patient-level canaries in GPT-2-style models trained on MIMIC-IV, across early, best-validation, and final checkpoints. Reconstruction improves with additional context, but training--held-out differences remain small. Trajectory-unique prompts containing limited clinical information produce no excess sensitive disclosure for training patients. In a matched model with random binary patient-level canaries, training-patient label-recovery AUROC increases from 0.512 to 0.549 while held-out recovery remains at chance. An overfit 1,000-patient positive control achieves near-perfect recovery, confirming sensitivity to strong memorisation. These results reveal weak cohort-level memorisation but no reliable patient-level disclosure under the evaluated attacks. They motivate clinically grounded inference-time audits that can detect patient-specific memorisation in existing models without retraining with canaries.
\end{abstract}

\section{Introduction}
Electronic health records (EHRs) contain longitudinal accounts of diagnoses, medications, procedures, laboratory measurements, and healthcare encounters. Their sensitivity and potential identifiability restrict access, limiting clinical research and model development. EHR foundation models learn from these records to support clinical forecasting, risk estimation, and synthetic-data generation \citep{kraljevic_2024_foresight,renc_2024_zero,fallahpour_2025_ehrmamba}.

Many such models represent patient histories as event sequences and predict their continuation. This autoregressive design creates an inference-time privacy concern: a partial history may prompt the model to reproduce sensitive events retained from training \citep{carlini_2021_extracting}. De-identification does not necessarily prevent disclosure because combinations of clinical events can still form distinctive trajectories.

Auditing this risk requires distinguishing memorisation from clinical generalisation. Correlated clinical events make accurate continuations compatible with either patient-specific recall or common patterns of care. Additional context improves prediction while potentially making training trajectories easier to recover. Aggregate privacy or validation metrics may also conceal leakage concentrated among a small number of patients.

We investigate how inference-time tests can probe memorisation across training: how the amount of patient information affects reconstruction, whether limited context elicits sensitive attributes, and whether clinically independent controls distinguish patient-specific retention from generalisation. Our aim is to develop practical, interpretable privacy audits for generative models of longitudinal health records.

\section{Related Work}

\noindent\textbf{EHR foundation models.} Foundation models for longitudinal EHRs represent patient histories as sequences of diagnoses, procedures, medications, laboratory results, and other clinical events. Models such as Foresight \citep{kraljevic_2024_foresight} and ETHOS \citep{renc_2024_zero} use autoregressive transformers to forecast future events or generate complete patient trajectories, while CEHR-GPT \citep{pang_2024_cehrgpt} extends this approach to representation learning and synthetic-record generation. 
Although these models support flexible clinical prediction, their ability to continue a partially observed patient trajectory creates a direct route through which sensitive patient records in the training data could be reconstructed at inference.

\noindent\textbf{Privacy and memorisation.} Most healthcare privacy evaluations have focused on synthetic EHR releases, using nearest-record distances and membership or attribute inference attacks to compare generated records with the training data \citep{yan_2022_multifaceted, yoon_2023_ehrsafe, zhang_2022_membership}. Related work on clinical-text language models has examined the extraction of patient identifiers and exact training-text reproduction \citep{lehman_2021_does, li_2026_memorization}, including how memorisation changes during training. However, considerably less work has examined autoregressive models of structured patient trajectories. The closest prior work proposes black-box tests for trajectory reconstruction, sensitive-attribute disclosure, membership inference, and patient-level risk in a structured EHR foundation model \citep{tonekaboni_2025_investigation}. In this setting, distinguishing patient-specific memorisation from clinical generalisation remains difficult because many patient trajectories contain common and predictable clinical patterns.

\noindent\textbf{Canary-based auditing.} Canaries are random artificial sequences inserted into training data to provide a known target for measuring memorisation. An exposure metric compares a canary's likelihood with alternative sequences drawn from the same space, with unusually high likelihood indicating retention of the inserted secret \citep{carlini_2019_secret}. Subsequent work has extended canary-based auditing to estimate privacy leakage more efficiently, including from a single training run \citep{steinke_2023_privacy}. These methods motivate the use of information independent of the original data as a controlled test of memorisation; we adapt this general principle to longitudinal EHR models.

\section{Methodology}

We train matched autoregressive models on 228,508 MIMIC-IV v2.2 \citep{johnson_2023_mimiciv, johnson_2023_mimiciva} patients using identical architectures and hyperparameters. Tokenisation was done via the ETHOS framework \citep{renc_2024_zero}; further details are provided in Appendix~\ref{app:loss}. One model is trained on the original patient trajectories, while the other includes the binary canaries described below. We evaluate three checkpoints representing distinct stages of training. Let $L_0$ be the initial validation loss and $L_{\mathrm{best}}$ the minimum validation loss. The early checkpoint is the first checkpoint satisfying
\[
L_{\mathrm{val}} \leq \frac{L_0 + L_{\mathrm{best}}}{2},
\]
the best checkpoint has the lowest validation loss, and the late checkpoint is the final training checkpoint when the allocated four-day training budget was exhausted. Training curves and computational details are provided in Appendix \ref{app:loss}.

We probe for memorisation using three complementary tests:

\noindent\textbf{Sequence reconstruction.} We select 3,000 training and 3,000 held-out patients with trajectories of at least 500 tokens, using the same patients at each checkpoint. We provide varying context lengths ending at token 400 and greedily reconstruct tokens 401--500. We measure the per-token reconstruction fraction across the 100-token target window and compare training with held-out patients. We use this fixed window framework to isolate the effect of the context length.

\noindent\textbf{Sensitive-attribute disclosure.} We select 300 training and 300 held-out patients whose trajectories contain a depression diagnosis, chosen as its relatively high dataset prevalence ($\sim12\%$) may facilitate more tractable extraction. For each patient, we provide the shortest prefix that uniquely identifies their trajectory relative to the training set. The depression token is absent from every prompt. From each prefix, we sample 20 continuations at temperature 1, generating up to 1,000 new tokens and stopping when either the depression diagnosis or the end-of-sequence token is produced. This minimal-context construction reduces interference from clinically predictive information, such as medications associated with depression. We report continuation-level disclosure rates—the fraction of continuations that generate the depression token—with 95\% confidence intervals computed across patients. As a negative control, we apply the same procedure to 300 training and 300 held-out patients whose complete trajectories do not contain the depression token, matched within each split to the depression-positive cohort on unique-prefix length and approximately on total trajectory length evaluated on the best checkpoint.

\noindent\textbf{Canary audit.} To distinguish memorisation from clinical generalisation, we train a separate canary-augmented model inspired by prior work on unintended memorisation \citep{carlini_2019_secret}. Each patient is randomly assigned a balanced binary canary, A or B, appended immediately before the end-of-sequence token. We measure how confidently the model recovers the assigned canary from the preceding clinical trajectory. We calculate a signed memorisation score $\mu_i$ measuring how strongly the model favours each patient’s assigned canary over the alternative. The \(\tau(\gamma)\) statistic measures the excess proportion of patients whose assigned canary is favoured above a specific threshold ($ \mu_i> \gamma$). Formal definitions are provided in Appendix \ref{app:canary}. Recovery above the held-out null indicates learning of arbitrary patient-specific associations that cannot be explained by clinical generalisation. We also evaluate a deliberately overfit model trained on just 1,000 patients (0.4\% of the original training dataset) as a positive control.

\section{Results}

\begin{figure}[H]
\centering
\includegraphics[width=0.8\textwidth]{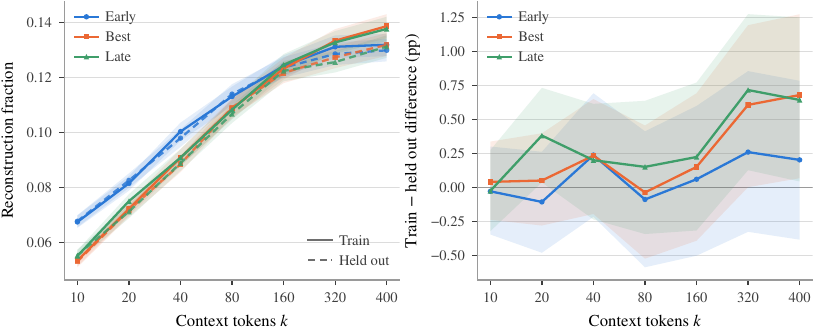}
\caption{Sequence reconstruction across training. Models greedily reconstruct tokens 401--500 from varying amounts of preceding context. Lines compare training and held-out patients; shaded regions show 95\% patient-bootstrap confidence intervals.}
\label{fig:recon}
\end{figure}

At checkpoints 30k, 550k, and 730k, reconstruction improved with context length in both training and held-out patients (Figure \ref{fig:recon}), so accuracy alone cannot distinguish memorisation from clinical generalisation. A small training–held-out gap of 0.6–0.7 percentage points emerged at the best and late checkpoints with 320–400 context tokens, consistent with limited training-data retention. 

Table~\ref{tab:disclosure} shows no evidence of excess depression disclosure for training patients. The training–held-out difference was small at every checkpoint (\(+0.58\), \(-0.47\), and \(-1.45\) percentage points), and all 95\% confidence intervals included zero. Thus, a trajectory-unique prefix was insufficient to recover patient-specific depression information under this attack. This null result does not exclude memorisation: the deliberately short prompts contain little clinically predictive context, while depression can appear much later in the trajectory and may fall beyond the generation window.

\begin{table}[t]
\centering
\caption{Random-canary recovery across training, including the overfit positive control. We report AUROC, the mean memorisation score ($\bar{\mu}$), and the excess-tail statistic $\tau(0.1)$. AUROC intervals are 95\% percentile bootstrap confidence intervals obtained from 10,000 patient-level resamples within each cohort.}
\label{tab:canary-auroc}
\small
\resizebox{\linewidth}{!}{%
\begin{tabular}{@{}l@{\hspace{0pt}}r lrr lrr@{}}
\toprule
& & \multicolumn{3}{c}{Train} & \multicolumn{3}{c}{Held-out} \\
\cmidrule(lr){3-5}\cmidrule(lr){6-8}
Checkpoint & Step & AUROC [95\% CI] & $\bar{\mu}$ & $\tau(0.1)$
           & AUROC [95\% CI] & $\bar{\mu}$ & $\tau(0.1)$ \\
\midrule
\multicolumn{8}{@{}l}{\textit{Full-scale model} ($228{,}508$ members)} \\
Early & 30{,}000
      & 0.512 [0.510, 0.515] & $+0.002$ & $+0.009$
      & 0.503 [0.496, 0.510] & $0.000$ & $+0.002$ \\
Best & 550{,}000
     & 0.549 [0.547, 0.551] & $+0.008$ & $+0.036$
     & 0.500 [0.493, 0.507] & $0.000$ & $+0.005$ \\
Late & 730{,}000
     & 0.548 [0.546, 0.551] & $+0.008$ & $+0.035$
     & 0.498 [0.491, 0.505] & $0.000$ & $+0.001$ \\
\addlinespace
\multicolumn{8}{@{}l}{\textit{Positive control: overfit model} ($1{,}000$ members)} \\
Overfit & 10{,}000
        & 1.000 [1.000, 1.000] & $+0.969$ & $+0.987$
        & 0.500 [0.493, 0.506] & $-0.002$ & $-0.002$ \\
\bottomrule
\end{tabular}%
}
\end{table}

Canary recovery increased during training: training-patient AUROC rose from 0.512 to 0.549, while \(\tau(0.1)\) increased from 0.009 to 0.036 (see Appendix \ref{counts} for tail counts), consistent with weak population-level memorisation but with few high-confidence examples. Held-out recovery remained near chance. The overfit positive control achieved near-perfect training-patient recovery while remaining at chance on held-out patients, confirming sensitivity to strong memorisation.

\begin{table}[H]
\centering
\caption{Sensitive-attribute disclosure. Continuations disclosing a depression diagnosis, out of 6{,}000 per arm (300 patients $\times$ 20 rollouts). Intervals are 95\% percentile bootstrap confidence intervals for the training-minus-held-out difference, obtained from 10,000 patient-level resamples within each cohort. The marginal prevalence of the  depression token was 12.3\% in the training set and 12.1\% in the validation set.}
\label{tab:disclosure}
\begin{tabular}{@{}lr rr rr@{}}
\toprule
Checkpoint & Step & Training & Held-out & Gap (pp) & 95\% CI \\
\midrule
\multicolumn{6}{@{}l}{\textit{Depression-token carriers}} \\
Early & 30{,}000
      & 932 (15.5\%) & 897 (14.9\%)
      & $+0.58$ & $[-1.43,\, +2.65]$ \\
Best & 550{,}000
     & 982 (16.4\%) & 1{,}010 (16.8\%)
     & $-0.47$ & $[-2.63,\, +1.62]$ \\
Late & 730{,}000
     & 958 (16.0\%) & 1{,}045 (17.4\%)
     & $-1.45$ & $[-3.58,\, +0.62]$ \\
\addlinespace
\multicolumn{6}{@{}l}{\textit{Matched negative patients: false disclosure}} \\
Best & 550{,}000
     & 658 (11.0\%) & 719 (12.0\%)
     & $-1.02$ & $[-2.68,\, +0.63]$ \\
\bottomrule
\end{tabular}
\end{table}


Because the audit requires a separately trained canary-augmented model, it does not directly establish the degree of memorisation in the base model. However, nearly identical losses and similar downstream-task performance (Appendix~\ref{app:task}) suggest limited disruption from canary insertion, supporting the use of the audit as an indicator of potential memorisation in the base model.

\section{Conclusion}

We evaluated inference-time memorisation in an autoregressive EHR foundation model across three stages of training. Trajectory reconstruction improved with additional context but differed little between training and held-out patients, while trajectory-unique prompts produced no excess depression disclosure. We found no evidence of reliable patient-specific reconstruction using the evaluated attacks. The canary audit provided the clearest evidence of memorisation beyond clinical generalisation, detecting weak cohort-level retention of randomly assigned patient-specific labels. However, this does not directly establish clinical-information leakage in the original model.

Future work should better distinguish memorisation from clinical generalisation, potentially building on approaches explored in language models \citep{morris_2026_how}. Longer training runs with alternative learning-rate schedules could clarify how overfitting relates to memorisation, membership inference vulnerability and the reconstruction of sensitive patient information. Systematically measuring canary memorisation across model capacity, dataset size and training duration could then identify higher- and lower-risk regimes and establish whether these findings transfer to models trained without canaries.

\clearpage

\begin{ack}

Inigo Stibbe and Joshua Placidi are supported by the UKRI AI Centre for Doctoral Training in Digital Healthcare (EP/Y030974/1). Joshua Placidi is also supported by a philanthropic Bupa PhD Scholarship. Padmanabhan Ramnarayan is supported by the UK National Institute for Health and Care Research (NIHR Research Professorship, NIHR306356). A.~Aldo Faisal acknowledges support from the UKRI AI programme and the Engineering and Physical Sciences Research Council (EPSRC) for the AI Hub in Generative Models (EP/Y028805/1).

\end{ack}

\bibliographystyle{plain}
\bibliography{ref_new}

@inproceedings{carlini_2019_secret,
  title = {The Secret Sharer: Evaluating and Testing Unintended Memorization in Neural Networks},
  booktitle = {Proceedings of the 28th {{USENIX}} Conference on Security Symposium},
  author = {Carlini, Nicholas and Liu, Chang and Erlingsson, {\'U}lfar and Kos, Jernej and Song, Dawn},
  year = 2019,
  series = {{{SEC}}'19},
  pages = {267--284},
  publisher = {USENIX Association},
  address = {USA},
  isbn = {978-1-939133-06-9}
}

@inproceedings{carlini_2021_extracting,
  title = {Extracting Training Data from Large Language Models},
  booktitle = {30th {{USENIX}} Security Symposium ({{USENIX}} Security 21)},
  author = {Carlini, Nicholas and Tram{\`e}r, Florian and Wallace, Eric and Jagielski, Matthew and {Herbert-Voss}, Ariel and Lee, Katherine and Roberts, Adam and Brown, Tom and Song, Dawn and Erlingsson, {\'U}lfar and Oprea, Alina and Raffel, Colin},
  year = 2021,
  month = aug,
  pages = {2633--2650},
  publisher = {USENIX Association},
  isbn = {978-1-939133-24-3}
}

@inproceedings{fallahpour_2025_ehrmamba,
  title = {{{EHRMamba}}: {{Towards}} Generalizable and Scalable Foundation Models for Electronic Health Records},
  booktitle = {Proceedings of the 4th Machine Learning for Health Symposium},
  author = {Fallahpour, Adibvafa and Alinoori, Mahshid and Ye, Wenqian and Cao, Xu and Afkanpour, Arash and Krishnan, Amrit},
  editor = {Hegselmann, Stefan and Zhou, Helen and Healey, Elizabeth and Chang, Trenton and Ellington, Caleb and Mhasawade, Vishwali and Tonekaboni, Sana and Argaw, Peniel and Zhang, Haoran},
  year = 2025,
  month = dec,
  series = {Proceedings of Machine Learning Research},
  volume = {259},
  pages = {291--307},
  publisher = {PMLR}
}

@article{johnson_2023_mimiciv,
  title = {{{MIMIC-IV}}, a Freely Accessible Electronic Health Record Dataset},
  author = {Johnson, Alistair E. W. and Bulgarelli, Lucas and Shen, Lu and Gayles, Alvin and Shammout, Ayad and Horng, Steven and Pollard, Tom J. and Hao, Sicheng and Moody, Benjamin and Gow, Brian and Lehman, Li-wei H. and Celi, Leo A. and Mark, Roger G.},
  year = 2023,
  month = jan,
  journal = {Scientific Data},
  volume = {10},
  number = {1},
  pages = {1},
  issn = {2052-4463},
  doi = {10.1038/s41597-022-01899-x}
}

@article{kraljevic_2024_foresight,
  title = {Foresight---a Generative Pretrained Transformer for Modelling of Patient Timelines Using Electronic Health Records: A Retrospective Modelling Study},
  author = {Kraljevic, Zeljko and Bean, Dan and Shek, Anthony and Bendayan, Rebecca and Hemingway, Harry and Yeung, Joshua Au and Deng, Alexander and Balston, Alfred and Ross, Jack and Idowu, Esther and Teo, James T and Dobson, Richard J B},
  year = 2024,
  month = apr,
  journal = {The Lancet Digital Health},
  volume = {6},
  number = {4},
  pages = {e281-e290},
  publisher = {Elsevier},
  issn = {2589-7500},
  doi = {10.1016/S2589-7500(24)00025-6},
  urldate = {2026-09-25}
}

@inproceedings{lehman_2021_does,
  title = {Does {{BERT}} Pretrained on Clinical Notes Reveal Sensitive Data?},
  booktitle = {Proceedings of the 2021 Conference of the {{North American}} Chapter of the {{Association}} for {{Computational Linguistics}}: {{Human}} Language Technologies},
  author = {Lehman, Eric and Jain, Sarthak and Pichotta, Karl and Goldberg, Yoav and Wallace, Byron C.},
  editor = {Toutanova, Kristina and Rumshisky, Anna and Zettlemoyer, Luke and {Hakkani-Tur}, Dilek and Beltagy, Iz and Bethard, Steven and Cotterell, Ryan and Chakraborty, Tanmoy and Zhou, Yichao},
  year = 2021,
  month = jun,
  pages = {946--959},
  publisher = {Association for Computational Linguistics},
  address = {Online},
  doi = {10.18653/v1/2021.naacl-main.73}
}

@article{li_2026_memorization,
  title = {Memorization in Large Language Models in Medicine Prevalence Characteristics and Implications},
  author = {Li, Anran and Qian, Lingfei and Du, Mengmeng and Yin, Yu and Hu, Yan and Sun, Zihao and Fu, Yihang and Kim, Hyunjae and Stutz, Erica and Ai, Xuguang and Xie, Qianqian and Zhu, Rui and Huang, Jimin and Yang, Yifan and Liu, Siru and Tham, Yih-Chung and {Ohno-Machado}, Lucila and Cho, Hyunghoon and Lu, Zhiyong and Xu, Hua and Chen, Qingyu},
  year = 2026,
  month = jun,
  journal = {Nature Communications},
  volume = {17},
  number = {1},
  pages = {7729},
  issn = {2041-1723},
  doi = {10.1038/s41467-026-73779-6}
}

@article{pang_2024_cehrgpt,
  title = {{{CEHR-GPT}}: {{Generating}} Electronic Health Records with Chronological Patient Timelines},
  author = {Pang, Chao and Jiang, Xinzhuo and Pavinkurve, Nishanth Parameshwar and Kalluri, Krishna S. and Minto, Elise L. and Patterson, Jason and Zhang, Linying and Hripcsak, George and G{\"u}rsoy, Gamze and Elhadad, No{\'e}mie and Natarajan, Karthik},
  year = 2024,
  journal = {arXiv preprint arXiv:2402.04400},
  eprint = {2402.04400},
  primaryclass = {cs.LG},
  doi = {10.48550/arXiv.2402.04400},
  archiveprefix = {arXiv}
}

@article{renc_2024_zero,
  title = {Zero Shot Health Trajectory Prediction Using Transformer},
  author = {Renc, Pawel and Jia, Yugang and Samir, Anthony E. and Was, Jaroslaw and Li, Quanzheng and Bates, David W. and Sitek, Arkadiusz},
  year = 2024,
  month = sep,
  journal = {npj Digital Medicine},
  volume = {7},
  number = {1},
  pages = {256},
  issn = {2398-6352},
  doi = {10.1038/s41746-024-01235-0}
}

@inproceedings{steinke_2023_privacy,
  title = {Privacy {{Auditing}} with {{One}} (1) {{Training Run}}},
  booktitle = {Thirty-Seventh {{Conference}} on {{Neural Information Processing Systems}}},
  author = {Steinke, Thomas and Nasr, Milad and Jagielski, Matthew},
  year = 2023,
  month = nov,
  urldate = {2026-09-07},
  langid = {english}
}

@inproceedings{tonekaboni_2025_investigation,
  title = {An Investigation of Memorization Risk in Healthcare Foundation Models},
  booktitle = {The Thirty-Ninth Annual Conference on Neural Information Processing Systems},
  author = {Tonekaboni, Sana and Stempfle, Lena and Fallahpour, Adibvafa and Gerych, Walter and Ghassemi, Marzyeh},
  year = 2025
}

@article{yan_2022_multifaceted,
  title = {A {{Multifaceted}} Benchmarking of Synthetic Electronic Health Record Generation Models},
  author = {Yan, Chao and Yan, Yao and Wan, Zhiyu and Zhang, Ziqi and Omberg, Larsson and Guinney, Justin and Mooney, Sean D. and Malin, Bradley A.},
  year = 2022,
  month = dec,
  journal = {Nature Communications},
  volume = {13},
  number = {1},
  pages = {7609},
  issn = {2041-1723},
  doi = {10.1038/s41467-022-35295-1}
}

@article{yoon_2023_ehrsafe,
  title = {{{EHR-Safe}}: Generating High-Fidelity and Privacy-Preserving Synthetic Electronic Health Records},
  author = {Yoon, Jinsung and Mizrahi, Michel and Ghalaty, Nahid Farhady and Jarvinen, Thomas and Ravi, Ashwin S. and Brune, Peter and Kong, Fanyu and Anderson, Dave and Lee, George and Meir, Arie and Bandukwala, Farhana and Kanal, Elli and Ar{\i}k, Sercan {\"O}. and Pfister, Tomas},
  year = 2023,
  month = aug,
  journal = {npj Digital Medicine},
  volume = {6},
  number = {1},
  pages = {141},
  issn = {2398-6352},
  doi = {10.1038/s41746-023-00888-7}
}

@article{zhang_2022_membership,
  title = {Membership Inference Attacks against Synthetic Health Data},
  author = {Zhang, Ziqi and Yan, Chao and Malin, Bradley A.},
  year = 2022,
  journal = {Journal of Biomedical Informatics},
  volume = {125},
  pages = {103977},
  issn = {1532-0464},
  doi = {10.1016/j.jbi.2021.103977}
}

@inproceedings{morris_2026_how,
  title = {How Much Can Language Models Memorize?},
  booktitle = {Forty-Third International Conference on Machine Learning},
  author = {Morris, John Xavier and Sitawarin, Chawin and Guo, Chuan and Kokhlikyan, Narine and Suh, G. Edward and Rush, Alexander M and Chaudhuri, Kamalika and Mahloujifar, Saeed},
  year = 2026
}

@article{johnson_2023_mimiciva,
  title = {{{MIMIC-IV}}},
  author = {Johnson, Alistair and Bulgarelli, Lucas and Pollard, Tom and Horng, Steven and Celi, Leo Anthony and Mark, Roger},
  year = 2023,
  month = jan,
  journal = {PhysioNet},
  doi = {10.13026/6mm1-ek67}
}
\clearpage
\appendix

\section{Data Representation and Training Details}
\label{app:loss}
\subsection{ETHOS Tokenisation}

ETHOS constructs Patient Health Timelines using a rule-based clinical tokenisation pipeline rather than a learned subword tokeniser. Events are extracted from 12 MIMIC-IV tables, standardised, and ordered chronologically. Medication identifiers are mapped to Anatomical Therapeutic Chemical (ATC) codes, while ICD-9 diagnoses are converted to ICD-10-CM where possible. Hierarchical ATC and ICD-10-CM codes are then divided into meaningful component tokens, whereas ICD-10-PCS procedure codes are tokenised character by character. Continuous measurements are mapped to deciles calculated from the training data. Finally, discrete time-interval tokens are inserted between events and the original timestamps are removed, producing a single chronological token sequence for each patient.

\begin{table}[htbp]
\centering
\small
\caption{\textbf{Dataset summary.} ETHOS-tokenised MIMIC-IV 2.2 patient trajectories.}
\label{tab:dataset}
\begin{tabular}{@{}lrrr@{}}
\toprule
& Train & Validation & Test \\
\midrule
Patients & 228{,}508 & 28{,}560 & 28{,}538 \\
Total tokens & 289{,}983{,}089 & 36{,}217{,}252 & 36{,}496{,}872 \\
\addlinespace
Length, mean $\pm$ sd
& $1{,}269 \pm 3{,}447$
& $1{,}268 \pm 3{,}252$
& $1{,}279 \pm 3{,}574$ \\
\quad minimum & 8 & 8 & 8 \\
\quad 25th percentile & 128 & 128 & 129 \\
\quad median & 337 & 341 & 346 \\
\quad 75th percentile & 1{,}059 & 1{,}062 & 1{,}069 \\
\quad maximum & 221{,}352 & 96{,}690 & 107{,}153 \\
\bottomrule
\end{tabular}
\end{table}

\subsection{Training Configurations}

\begin{table}[H]
\centering
\small
\caption{\textbf{Training Configuration of Base Model.} A GPT-2 decoder with rotary
position embeddings, trained on ETHOS-tokenised MIMIC-IV 2.2 trajectories. The
learning-rate schedule completes at step 500{,}000, so the remaining steps ran at the
floor rate of $6\times10^{-5}$.}
\label{tab:model-config}
\begin{tabular}{@{}lr@{}}
\toprule
\multicolumn{2}{@{}l}{\textit{Architecture}} \\
Layers & 6 \\
Attention heads & 12 \\
Embedding dimension & 768 \\
Context window & 2{,}048 \\
Positional encoding & RoPE ($10^4$) \\
Dropout & 0.1 \\
Vocabulary & 4{,}509 \\
Parameters & 49.5M \\
\addlinespace
\multicolumn{2}{@{}l}{\textit{Optimisation}} \\
Optimiser & AdamW \\
Learning rate & $6 \times 10^{-4}$ \\
$\beta_1$, $\beta_2$ & 0.9, 0.95 \\
Weight decay & 0.1 \\
Batch size & 128 \\
Training steps & 734{,}000 \\
\quad warmup / hold / decay & 4{,}000 / 136{,}000 / 360{,}000 \\
Final LR ratio & 0.1 \\
\bottomrule
\end{tabular}
\end{table}

\begin{figure}[H]
    \centering
    \includegraphics[width=0.75\linewidth]{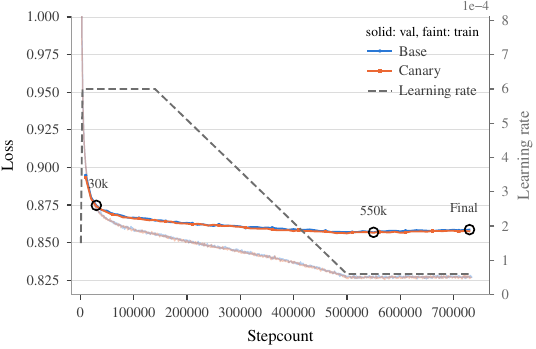}
    \caption{\textbf{Training and validation loss for the base and canary-augmented models.} Checkpoints are evaluated at 30k steps (early), 550k steps (best validation loss), and 730k steps (final). The early checkpoint is the first to achieve half of the total validation-loss improvement.}
    \label{fig:training-loss}
\end{figure}

\subsection{Overfit Reconstruction Test}

\begin{figure}[H]
    \centering
    \includegraphics[width=0.5\linewidth]{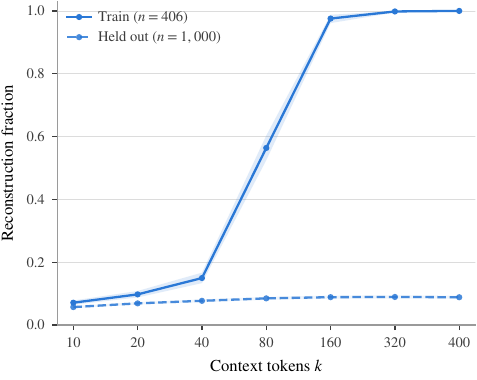}
    \caption{\textbf{Sequence-reconstruction positive control.} Only 406 of the 1000 train patients had sequence length greater than 500 tokens.}
    \label{fig:extra-1}
\end{figure}

\subsection{Sensitive-Attribute Audit Cohort}

\begin{figure}[H]
    \centering
    \includegraphics[width=0.5\linewidth]{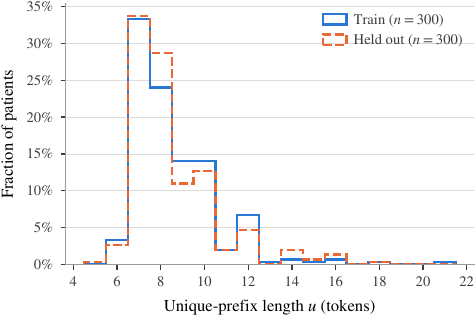}
    \caption{\textbf{Prompt-length distributions for sensitive-attribute disclosure.}
    Shortest unique-prefix lengths for the 300 training and 300 held-out validation patients selected for the depression disclosure audit. Histograms show the fraction of patients at each prefix length.}    
    \label{fig:extra-2}
\end{figure}

\begin{figure}[H]
    \centering
    \includegraphics[width=0.5\linewidth]{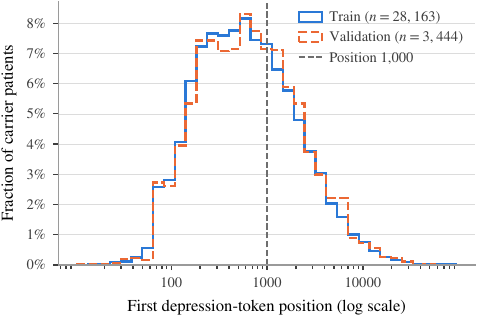}
    \caption{First depression-token positions among training and validation carriers. The dashed line marks absolute position 1,000; 66.3\% and 64.3\% of carriers, respectively, have their first occurrence at or before this position }
    \label{fig:extra-3}
\end{figure}

\section{Canary-Based Metric Derivations}
\label{app:canary}

\subsection{The Memorisation Score: $\mu_i$}

Within each split, we randomly permute a balanced collection of binary labels $l_i \in \{A,B\}$ across patient trajectories $u_i$, appending each assigned label to the corresponding trajectory, such that
\[
P(l_i=A)=P(l_i=B)=\tfrac{1}{2}.
\]
The labels are therefore independent of patient content, although assignments are not mutually independent because their totals are fixed. Systematic recovery of these arbitrary associations provides evidence of patient-specific memorisation, but does not itself demonstrate extraction of clinical events.

Let $z_{i,A}$ and $z_{i,B}$ denote the model logits for the two canary tokens after conditioning on $u_i$. We define the pairwise probability assigned to the true canary as
\[
\pi_i =
\begin{cases}
\sigma(z_{i,A}-z_{i,B}), & l_i=A,\\
\sigma(z_{i,B}-z_{i,A}), & l_i=B,
\end{cases}
\]
where $\sigma$ is the logistic sigmoid. Under the random-assignment null,
$\mathbb{E}_{l_i}[\pi_i \mid u_i]=\tfrac{1}{2}$, although individual values may differ from $\tfrac{1}{2}$.

We define the patient-level memorisation score
\begin{equation}
    \mu_i=2\pi_i-1,
\end{equation}
where $\mu_i=0$ indicates no preference between the canaries, positive values favour the assigned canary, and negative values favour the alternative. Under the null, the distribution of $\mu_i$ is expected to be approximately symmetric about zero. We therefore define an excess-tail statistic $\tau(\gamma)$ measuring whether strongly positive scores occur more frequently than equally strong negative scores.

\subsection{Excess-Tail Memorisation Statistic: $\tau(\gamma)$}

\label{app:metric}

For a threshold $\gamma \in [0,1]$, we define the positive- and negative-tail
rates as
\begin{equation}
    T^{+}(\gamma)
    =
    \frac{1}{N}\sum_{i=1}^{N}
    \mathbb{I}\{\mu_i > \gamma\},
    \qquad
    T^{-}(\gamma)
    =
    \frac{1}{N}\sum_{i=1}^{N}
    \mathbb{I}\{\mu_i < -\gamma\}.
\end{equation}

Because the canary labels are assigned independently of the patient
trajectories, the distribution of $\mu_i$ should be approximately symmetric
about zero when the patient--canary associations have not been memorised.
We therefore define the excess-tail memorisation statistic as
\begin{equation}
    \tau(\gamma)
    =
    T^{+}(\gamma)-T^{-}(\gamma)
    =
    \frac{1}{N}\sum_{i=1}^{N}
    \left[
        \mathbb{I}\{\mu_i>\gamma\}
        -
        \mathbb{I}\{\mu_i<-\gamma\}
    \right].
    \label{eq:tau}
\end{equation}

Positive values of $\tau(\gamma)$ indicate an excess of patients for whom
the model strongly favours the assigned canary over patients for whom it
equally strongly favours the alternative canary. Under the no-memorisation
null, we expect $\tau(\gamma)$ to be close to zero.

\subsection{Positive and Negative Canary-Score Tail Counts}
\label{counts}

\begin{table}[htbp]
\centering
\caption{Separate canary-score tail
counts. Each entry lists $N_+\,/\,N_-$: the numbers of patients
with $\mu_i>\gamma$ and $\mu_i<-\gamma$, respectively.
The slash separates counts and does not denote a ratio.
Full-scale cohorts contain 228,508 training and 28,560 held-out
validation patients; the overfit cohort contains 1,000 training
and 28,560 held-out validation patients.}
\label{tab:canary-tails}

\small
\begin{tabular}{@{}lrrrr@{}}
\toprule
& \multicolumn{2}{c}{$\gamma=0.1$}
& \multicolumn{2}{c}{$\gamma=0.5$} \\
\cmidrule(lr){2-3}\cmidrule(lr){4-5}
Checkpoint & Train & Held out & Train & Held out \\
\midrule
Early
& $39{,}631\,/\,37{,}508$ & $4{,}783\,/\,4{,}724$
& $0\,/\,0$ & $0\,/\,0$ \\
Best
& $30{,}529\,/\,22{,}311$ & $3{,}354\,/\,3{,}224$
& $1\,/\,1$ & $0\,/\,1$ \\
Late
& $32{,}017\,/\,23{,}952$ & $3{,}535\,/\,3{,}497$
& $0\,/\,1$ & $0\,/\,0$ \\
\addlinespace
Overfit
& $991\,/\,4$ & $13{,}746\,/\,13{,}792$
& $979\,/\,2$ & $11{,}536\,/\,11{,}588$ \\
\bottomrule
\end{tabular}
\end{table}


\section{Downstream Task Results}
\label{app:task}

Downstream performance is evaluated on the test set using a rollout-based benchmark framework. For each task, we identify a prediction point and two task-specific outcome tokens within the patient trajectory. The model is given the history preceding the prediction point and generates autoregressively until it produces one of the outcome tokens or reaches the maximum continuation length of 4,096 tokens. For the hospital readmission task specifically, a rollout is positive if a hospital admission token occurs within 30 days and negative if mortality occurs first or the 30-day horizon is exceeded. We sample 20 stochastic continuations per prompt and use the proportion terminating with the positive outcome token as the score for binary classification. Rollouts that do not reach either outcome within 4,096 tokens are excluded. We then calculate AUROC and AUPRC from the resulting prompt-level scores. Table~\ref{tab:benchmark_definitions} provides the prediction points and outcome tokens used for each task. Because rollout-based evaluation is stochastic and computationally expensive, we were unable to repeat the complete evaluation across multiple random seeds.

\begin{table}[H]
\centering
\caption{Benchmark definitions used in the rollout-based evaluation framework.}
\label{tab:benchmark_definitions}
\small
\begin{tabular}{@{}lll@{}}
        \toprule
        \textbf{Benchmark} & \textbf{Prediction point} & \textbf{Terminating tokens} \\
        \midrule
        ICU Mortality
        & \texttt{ICU\_ADMISSION}
        & \texttt{MEDS\_DEATH}, \texttt{ICU\_DISCHARGE} \\

        ICU Readmission
        & \texttt{ICU\_DISCHARGE}
        & \texttt{ICU\_ADMISSION}, \texttt{HOSPITAL\_DISCHARGE} \\

        ICU Admission
        & \texttt{HOSPITAL\_ADMISSION}
        & \texttt{ICU\_ADMISSION}, \texttt{HOSPITAL\_DISCHARGE} \\

        Hospital Mortality
        & \texttt{HOSPITAL\_ADMISSION}
        & \texttt{MEDS\_DEATH}, \texttt{HOSPITAL\_DISCHARGE} \\

        Hospital Readmission
        & \texttt{HOSPITAL\_DISCHARGE}
        & \texttt{HOSPITAL\_ADMISSION}, \texttt{MEDS\_DEATH} \\
        \bottomrule
    \end{tabular}
\end{table}

\begin{table}[H]
\centering
\caption{Downstream performance of the best base and canary-augmented models. Canary insertion produced little change in aggregate predictive performance.}
\label{tab:canary_downstream}
\small
\begin{tabular}{lcccc}
        \toprule
        & \multicolumn{2}{c}{AUROC} & \multicolumn{2}{c}{AUPRC} \\
        \cmidrule(lr){2-3}\cmidrule(lr){4-5}
        Task & Base & Canary & Base & Canary \\
        \midrule
        Hospital mortality   & 0.858 & 0.848 & 0.318 & 0.297 \\
        Hospital readmission & 0.703 & 0.703 & 0.486 & 0.482 \\
        ICU admission        & 0.912 & 0.911 & 0.767 & 0.766 \\
        ICU mortality        & 0.878 & 0.887 & 0.447 & 0.486 \\
        ICU readmission      & 0.735 & 0.724 & 0.400 & 0.388 \\
        \midrule
        Macro average        & 0.817 & 0.815 & 0.484 & 0.484 \\
        \bottomrule
    \end{tabular}
\end{table}
\end{document}